\documentclass[conference]{IEEEtran}
\usepackage{booktabs}
\usepackage{amsmath,amsthm,amssymb,amsfonts}
\usepackage{pgfplots}
\usepackage{hyperref}

\usepackage{mathtools}
\newcommand{\argmaxm}[1]{%
  \ifthenelse{\isempty{#1}}%
    {\overset{m}{\argmax}}
    {\underset{#1}{\overset{m}{\argmax}}\, }
}

\newcommand{\argminm}[1]{%
  \ifthenelse{\isempty{#1}}%
    {\overset{m}{\argmin}}
    {\underset{#1}{\overset{m}{\argmin}}\, }
}
\newcommand{\argmaxmset}[1]{%
  \ifthenelse{\isempty{#1}}%
    {\overset{[m]}{\argmax}}
    {\underset{#1}{\overset{[m]}{\argmax}}\, }
}
\newcommand{\argminmset}[1]{%
  \ifthenelse{\isempty{#1}}%
    {\overset{[m]}{\argmin}}
    {\underset{#1}{\overset{[m]}{\argmin}}\, }
}
 \DeclareMathOperator*{\argmax}{arg\,max}
 \DeclareMathOperator*{\argmin}{arg\,min}
\usepackage{algorithmic}
\pgfplotsset{compat=1.15}
\usepgfplotslibrary{
  statistics,
  colorbrewer,
  groupplots,
}
\usetikzlibrary{
  patterns,
  shapes.geometric,
  decorations.text,
  matrix,
  fit,
  backgrounds,
  positioning,
}
\tikzset{
  fignode/.style={
    outer sep=0.25em,
  }
}
\tikzset{
  framedfignode/.style={
    outer sep=0.25em,
    inner sep=0.5em,
    rounded corners,
    draw,
  }
}
\colorlet{plotColorNeutral}{gray}
\definecolor{plotColor1}{HTML}{f61a1c}
\definecolor{plotColor2}{HTML}{377eb8}
\definecolor{plotColor3}{HTML}{4daf4a}
\definecolor{plotColor4}{HTML}{984ea3}
\definecolor{plotColor5}{HTML}{FFFFCB}
\definecolor{plotColor6}{HTML}{1e90ff}
\colorlet{plotColorNeutral*}{plotColorNeutral!40}
\colorlet{plotColor1*}{plotColor1!60}
\colorlet{plotColor2*}{plotColor2!60}
\colorlet{plotColor3*}{plotColor3!60}
\colorlet{plotColor4*}{plotColor4!60}
\colorlet{plotColor5*}{plotColor5!60}
\colorlet{plotColor6*}{plotColor6!60}
\pgfplotsset{
    colormap={greenred}{HTML=(4daf4a) HTML=(e41a1c)},
    colormap={redgreen}{HTML=(e41a1c) HTML=(4daf4a)}
}
\usepackage{graphicx}
\usepackage{graphicx}
\usepackage{colortbl}
\definecolor{blue}{RGB}{17,220,247}
\definecolor{purple}{RGB}{163,115,250}
\definecolor{caribbeangreen}{rgb}{0.0, 0.8, 0.6}

\definecolor{bluecolor}{RGB}{0,0,255}

\definecolor{GREEN}{RGB}{84,130,53}

\usepackage{wrapfig}

\usepackage{bbm}
\graphicspath{ {imgs/} }
\usepackage{svg}
\newcommand{\NA}[2]{N_{#1}(#2)}

\usepackage{enumitem}
\usepackage{subcaption}
\usepackage{cite}
\usepackage{algorithm}
\makeatletter
\newcommand\fs@betterruled{%
  \def\@fs@cfont{\bfseries}\let\@fs@capt\floatc@ruled
  \def\@fs@pre{\vspace*{5pt}\hrule height.8pt depth0pt \kern2pt}%
  \def\@fs@post{\kern2pt\hrule\relax}%
  \def\@fs@mid{\kern2pt\hrule\kern2pt}%
  \let\@fs@iftopcapt\iftrue}
\floatstyle{betterruled}
\restylefloat{algorithm}
\makeatother
\newtheorem{lemma}{Lemma}
\newtheorem{theorem}{Theorem}
\newcommand{\ARMS}{[K]}
\newcommand{\name}{CCS}

\newcommand{\EMPTOPTAU}{\mbox{$\hat{S}^{\tau_{\delta}}_m$}}
\newcommand{\EMPWORSTTAU}[1]{(\EMPTOPTAU)^c}

\newcommand{\HA}[1]{\text{H}^{\varepsilon}(\mu)}

\newtheorem{definition}{Definition}

\begin{document}

\title{Challenger-Based Combinatorial Bandits for Subcarrier Selection in OFDM Systems}

\author{
\IEEEauthorblockN{Mohsen Amiri, Venktesh V, and Sindri Magnússon}
\IEEEauthorblockA{Department of Computer and Systems Sciences (DSV)\\
Stockholm Univeristy\\
Email: \{mohsen.amiri, venktesh.viswanathan, sindri.magnusson\}@dsv.su.se}
}

\maketitle

\begin{abstract}
This paper investigates the identification of the top-m user-scheduling sets in multi-user MIMO downlink, which is cast as a combinatorial pure-exploration problem in stochastic linear bandits. Because the action space grows exponentially, exhaustive search is infeasible. We therefore adopt a linear utility model to enable efficient exploration and reliable selection of promising user subsets. We introduce a gap-index framework that maintains a shortlist of current estimates of champion arms (top-m sets) and a rotating shortlist of challenger arms that pose the greatest threat to the champions. This design focuses on measurements that yield the most informative gap-index-based comparisons, resulting in significant reductions in runtime and computation compared to state-of-the-art linear bandit methods, with high identification accuracy. The method also exposes a tunable trade-off between speed and accuracy. Simulations on a realistic OFDM downlink show that shortlist-driven pure exploration makes online, measurement-efficient subcarrier selection practical for AI-enabled communication systems.
\end{abstract}

\section{Introduction}
Orthogonal frequency-division multiplexing (OFDM) is the physical-layer workhorse of LTE, 5G NR, and emerging 6G systems, where a key scheduling problem is selecting which subcarriers to activate for each user or user group \cite{dahlman20205g}. Efficient selection is critical for spectral efficiency, user fairness, and diverse QoS targets, yet in each interval, the base station (BS) can only activate a fixed number \(m\) of subcarriers from a much larger pool of \(K\) due to hardware limits and policy constraints \cite{girici2008practical, busacca2025distributed}. Because per-subcarrier rates hinge on instantaneous, frequency-selective and thus time-varying channels, the quality of any size-\(m\) allocation fluctuates over time \cite{eldar2022machine, goldsmith2005wireless}. Exhaustively evaluating all \(\binom{K}{m}\) subsets is infeasible even at moderate scales, and each additional probe consumes valuable time–frequency resources \cite{huang2010subcarrier}. These realities motivate exploration strategies that are simultaneously \emph{sample-efficient} (minimizing costly measurements) and \emph{compute-efficient} (keeping per-round latency low).

Greedy heuristics and classical optimization work well at small scales but degrade or become intractable as \(K\) grows and channel measurements are noisy \cite{huang2010subcarrier, liu2008low, liu2013complexity}. In OFDM specifically, water-filling and optimization-based schedulers have a long track record and perform strongly in small systems, yet they scale poorly and typically assume full channel state information (CSI) and tractable oracles \cite{shams2014survey, wong1999multiuser, seong2006optimal, shen2020low}. Complexity-aware heuristics and clustering can cut computational cost but incur performance gaps under noisy observations \cite{huang2010subcarrier, liu2008low, liu2013complexity}, while recent ML/RL approaches improve adaptivity at the expense of substantial training and computing \cite{ye2024artificial}.

AI use in communication systems is accelerating, especially at the wireless physical layer (AI4PHY), enabling end-to-end transceiver design; meanwhile, 3GPP Rel-18/19 is beginning to standardize AI-based CSI feedback, beam management, and positioning \cite{ye2024artificial, eldar2022machine}. Within this trend, reinforcement learning (RL) is compelling, and bandit formulations are particularly attractive because they explicitly handle uncertainty. However, existing combinatorial and linear bandit methods either target only best-arm identification (\(m{=}1\)) \cite{xu2018fully} or depend on repeated, broad index recomputations across the action space that inflate per-round cost \cite{fiez2019sequentialexperimentaldesigntransductive, réda2021topm, reda2021top, chen2013combinatorial}.

Multi-armed bandits, especially linear and combinatorial variants, offer strong sample efficiency for best-arm identification and combinatorial pure exploration (CPE) \cite{chen2013combinatorial, fiez2019sequentialexperimentaldesigntransductive, xu2017fullyadaptivealgorithmpure, pmlr-v119-degenne20a, réda2021topm, reda2021top}. Yet, many approaches still require global index recomputation or wide action-space scans, which increases per-round latency and undermines millisecond-scale OFDM scheduling. Our work follows this bandit line but targets online top-\(m\) selection under \emph{low-dimensional}, \emph{noisy features} and strict latency, drawing shortlist inspiration from \cite{purohit2025sample} while avoiding global scans via a champion--challenger design. For OFDM schedulers operating on millisecond timescales, such overhead is prohibitive \cite{deniz2018performance}.



We model subcarrier selection as a CPE problem in stochastic linear bandits and introduce Champion--Challenger Sampling (\name{}): a latency-aware scheme that keeps a small set of current ``champions'' and a shortlist of ``challengers'' to query only the most informative comparisons instead of scanning all arms. The additive OFDM utility lets us solve subset selection at the arm level, and we link PHY measurements to the bandit model via a simple linear surrogate of per-subcarrier rate from noisy SNR. Compared to offline top-\(m\) selection like \cite{purohit2025sample}, which assumes high-dimensional, near-noiseless features and static ranking, our setting is online, uses low-dimensional noisy features, and explicitly trades accuracy for runtime by tuning the shortlist size. We provide \((\varepsilon,m,\delta)\)-PAC guarantees with instance-dependent complexity and empirically reduce arm pulls, comparisons, and runtime, achieving speedups of \textbf{100$\times$}, \textbf{15$\times$}, and \textbf{300$\times$} over LinGapE~\cite{xu2017fullyadaptivealgorithmpure}, LinUGapE~\cite{LinUGapE}, and LinGIFA~\cite{reda2021top}, respectively, while maintaining near-perfect top-\(m\) identification accuracy.

Overall, \name{} delivers a practical path to fast, online, reliable, and measurement-efficient subcarrier selection in OFDM systems: it targets only the comparisons that matter, comes with finite-sample guarantees, and empirically outperforms state-of-the-art linear-bandit baselines by large margins in both comparisons and runtime, which makes it applicable in AI-based communication systems.

\section{System Model and Problem Formulation}\label{sec:system_model}

We consider an OFDM downlink with \(K\) subcarriers \(\mathcal{C}=\{1,\dots,K\}\). In each scheduling interval, the transmitter assigns \(m\) subcarriers to one user or group; any candidate allocation is \(\mathcal{S}\subseteq\mathcal{C}\) with \(|\mathcal{S}|=m\), and the feasible set is \(\mathcal{A}=\{\,\mathcal{S}\subseteq\mathcal{C}\mid|\mathcal{S}|=m\,\}\).

Let $h_i$ denote the channel gain on subcarrier $i \in \mathcal{C}$. Assuming equal power allocation across active subcarriers and noise variance $\sigma^2$, the instantaneous signal-to-noise ratio (SNR) on subcarrier $i$ is
\begin{equation}
    \gamma_i \;=\; \frac{|h_i|^2 P_{\text{per-tone}}}{\sigma^2},
\end{equation}
where $P_{\text{per-tone}}$ is the transmit power per subcarrier.

The achievable rate on subcarrier $i$ is approximated as
\begin{equation} \label{eq:reward_noise_free}
    R_i \;=\; \log_2\!\left(1+\gamma_i\right).
\end{equation}
In practice, $\gamma_i$ is not observed exactly; instead we form a pilot-based estimate $\widehat{\gamma}_{i,t}$. Hence, even with a fixed channel within a slot, the measured rate is noisy. This motivates modeling the observed per-tone reward as the true mean rate plus an additive noise term. Therefore, the observed achievable rate on subcarrier $i$ with noisy SNR can be modeled as
\begin{equation} \label{eq:reward_noisey}
    \widehat{R}_i \;=\; \log_2\!\left(1+\gamma_i\right) + \eta_t,
\end{equation}
where $\eta_t$ is an additive noise.

\begin{lemma}[Sub-Gaussian reward noise under dB-SNR errors]
If the SNR estimation error in dB, $\xi_{i,t}$, is sub-Gaussian with proxy variance $\sigma_\xi^2$, then the induced reward noise
\[
\eta_t \;:=\; \log_2\!\big(1+\gamma_i\,10^{\xi_{i,t}/10}\big) - \log_2(1+\gamma_i)
\]
is sub-Gaussian with proxy variance $\sigma_\eta^2 \;\le\; \Big(\tfrac{\ln 10}{10\ln 2}\Big)^{\!2}\sigma_\xi^2$,
uniformly over $\gamma_i \ge 0$.
\end{lemma}

\begin{proof}
A convenient, realistic assumption is additive dB errors:
\begin{align} \label{eq:noisy_SNR}
\widehat{\gamma}^{\mathrm{dB}}_{i,t} \;=\; \gamma^{\mathrm{dB}}_i + \xi_{i,t}
\quad \Longleftrightarrow \quad
\widehat{\gamma}_{i,t} \;=\; \gamma_i \, 10^{\xi_{i,t}/10},
\end{align}
where $\xi_{i,t}$ is zero-mean sub-Gaussian with proxy variance $\sigma_\xi^2$ (Gaussian dB errors are common). The induced reward noise is \(\eta_t := \log_2(1+\gamma_i\,10^{\xi_{i,t}/10}) - \log_2(1+\gamma_i)\). Let \(g(\xi)=\log_2(1+\gamma_i\,10^{\xi/10})\); then \(g'(\xi)=\frac{\ln 10}{10\ln 2}\cdot\frac{\gamma_i\,10^{\xi/10}}{1+\gamma_i\,10^{\xi/10}}\le L_{\mathrm{dB}}\coloneqq \frac{\ln 10}{10\ln 2}\approx 0.332\), so by the mean-value theorem \(|\eta_t|=|g(\xi_{i,t})-g(0)|\le L_{\mathrm{dB}}\,|\xi_{i,t}|\). Because Lipschitz transformations preserve sub-Gaussian tails up to the constant, \(\eta_t\) is sub-Gaussian with proxy variance \(\sigma_\eta^2\le\big(\tfrac{\ln 10}{10\ln 2}\big)^{2}\sigma_\xi^2\).
\end{proof}

The utility of a subcarrier subset $\mathcal{S}$ is defined as the sum-rate across all allocated subcarriers:
\begin{equation}
    \phi(\mathcal{S}) = \sum_{i \in \mathcal{S}} \hat R_i.
\end{equation}
Our objective is to identify the \emph{top-$m$} channel subset $\mathcal{S}_\star^{(m)} \subseteq \mathcal{A}$ that maximize the expected utility. Let $\mu(\mathcal{S}) = \mathbb{E}[\phi(\mathcal{S})]$ denote the expected utility of set $\mathcal{S}$. Formally,
\begin{equation}
    \mathcal{S}_\star^{(m)} = \operatorname*{argmax}_{\mathcal{S} \in \mathcal{A}} \mu(\mathcal{S}).
\end{equation}

The action space has cardinality \( |\mathcal{A}| = \binom{K}{m} \), which grows exponentially in \(K\). Exhaustively evaluating all allocations is therefore infeasible in practice \cite{huang2010subcarrier, liu2008low, liu2013complexity}.

To enable tractable exploration, we adopt a \emph{linear utility model} \cite{purohit2025sample, réda2021topm, xu2018fully} and cast the problem as a top-$m$ arm selection in a multi-armed bandit setup: each subcarrier is an arm, and pulling an arm yields the noisy reward in Eq.~\ref{eq:reward_noisey}. 

\begin{equation}
    \mu(a) = \mathbf{x}_a^\top \boldsymbol{\theta}_\star,
\end{equation}
where $\mathbf{x}_a$ is the feature vector and $\boldsymbol{\theta}_\star \in \mathbb{R}^d$ is an unknown parameter.
\begin{algorithm}[!t]
 \small
 \caption{\strut \name{}}
 \label{algo:static_subsets}
 \small

  \begin{algorithmic}[1]
   \small

 \STATE \textbf{Input:}  $\mathcal{C}:$ set of subcarriers, $a\in \mathcal{C}$: an arm 
 \STATE \textbf{Define:} \hspace{1.5mm} $ U_t$: set of currently estimated top-$m$ arms, $C_t$: set of currently estimated next best-$m'$ arms, $b_t$: the most ambiguous arm from $U_t$, $ca_t$: the most ambiguous sampled arm from $C_t$, $U_0 \leftarrow$ set of random $m$ arms from $\mathcal{C}$,  $t \leftarrow 1$, \hspace{1mm} $\Vec{\theta}_{1} \leftarrow \mathcal{N}(0,1)$

 \small
\WHILE{$\neg$ ($B_t(ca_t,b_t) \le \epsilon$) }

            \STATE \textit{  Construct $U_t$ by replacing $n_t$ with better arm $c_t$} \\
            \STATE $n_{t} = \argmin_{a\in U_{t-1}} \hat{\mu}_{t}(a), ~~ c_{t} = \argmax_{a\in C_{t-1}} \hat{\mu}_{t}(a) $\\
           \IF{
           $\hat\mu_{t}(c_{t}) \ge \hat\mu_{t}(n_{t}) $ }
           
            \STATE $U_t,C_t \gets \mbox{swap}(n_t,c_t)$ from $U_{t-1},C_{t-1}$         
           \ENDIF

        \STATE $I_t \leftarrow s'\sim_{m'} (U_t\cup C_{t-1})^c$ // \textit{Randomly sample}
         
        \STATE $C_t \hspace{-1mm}\leftarrow\hspace{-1mm} \mbox{top}_{m'}(I_t \hspace{-0.5mm}\cup \hspace{-0.5mm}C_{t-1}; \hat{\mu}_{(t)})$  //\textit{Update $C_t$ from $C_{t-1}$ \hspace{-0.5mm}\&\hspace{-0.5mm} $I_t$}

        \STATE  \textsc{ Recompute Ambiguious arms} \\
        \STATE $b_{t+1} = \argmax_{b\in U_t} \max_{a\in C_t} \left[ B_t(a,b) \right]  $ \\
        \STATE $ca_{t+1} =  \argmax_{s\in C_t} \left[ B_t(s,b_{t+1}) \right]  $

        \STATE \textit{Pull selected arm, receive reward, and update parameters} \\
        \STATE $a_{t+1} \leftarrow$ selection\_rule$(U_t,C_t)$
        
        \STATE $r_{t+1} = R_t(a) + \eta_t$ \textit{ // Achievable rate  (observed reward)}

        \STATE $\hat{V}_{t+1} \hspace{-1mm} = \hspace{-1mm} \lambda I_n + \sum_{a\in\mathcal{S}} N_{a}\mathbf{x}_{a} \mathbf{x}_{a}^T$ //\textit{$\lambda$ regularized design matrix}\\
        
         \STATE $\hat{\theta}_{t+1}=(\hat{V}_{t+1})^{-1}(\sum_{l=1}^{t+1} r_l \mathbf{x}_{a_l})$ // \textit{Least-squares estimate}\\
         
        \STATE $t \gets t+1$
    \ENDWHILE
   
 \STATE \textbf{Output:}
 $U_T$: Set of $m$ arms which have the highest reward 
\end{algorithmic}
\end{algorithm}
In our setting, only a noisy estimate of the SNR for each subcarrier (arm) is observed yielding a noisy reward observation which is modeled as:
\begin{equation}
    \hat{\mu}_t(a_t) = \mathbf{x}_{a_t}^\top \boldsymbol{\hat\theta} + \eta_t,
\end{equation}
Our goal is to estimate $\hat\theta$ such that $\hat{\mu}(a_t) \approx  \mu(a)$

 Prior work shows that if a linear surrogate of the reward exists, top-\(m\) identification becomes possible via multi-armed bandit methods \cite{purohit2025sample, réda2021topm, xu2018fully}. Accordingly, we approximate the noise-free reward in Eq.~\ref{eq:reward_noise_free} for each arm \(i\) with a linear surrogate in terms of functions of the observed (noisy) SNR. Because the reward is the concave logarithm function, an \(N\)-term Taylor expansion yields
\begin{align}
\label{eq:reward_lin_surrogate}
R_i=\log_2\!\bigl(1+\gamma_i\bigr)
\approx \alpha_0 + \alpha_1 f_1(\gamma_i) + \cdots + \alpha_N f_N(\gamma_i)
\end{align}
where \(F_j = f_j(\gamma_i) = \gamma_i^j\) are the normalized features and \(\mathbf{x}_i = [F_1, F_2, \ldots, F_N]\) is the normalized feature vector. Since only the noisy SNR \(\hat{\gamma}_i\) from Eq.~\ref{eq:noisy_SNR} is observed, these normalized features are noisy as well. Nevertheless, because the noise is multiplicative, the surrogate in Eq.~\eqref{eq:reward_lin_surrogate} remains linear in the true SNR, so it suffices to compute the normalized features using the observed noisy SNR.





\section{Approach: Challenger Champion Sampling}
To address the above-mentioned limitations, we develop a sample and compute-efficient exploration strategy named \name{} that focuses only on the most promising candidates \cite{purohit2025sample}. Specifically, we introduce a challenger-based selective exploration algorithm, which maintains a small set of ``champion'' arms/subcarriers ($U_t$) along with a rotating shortlist of ``challenger'' arms ($C_t$). By focusing on the most informative arm comparisons between champions and a shortlist of challengers, the proposed approach significantly reduces the number of required computations while still guaranteeing reliable identification of the top-$m$ subcarriers.

An overview of the algorithm in as shown in Algorithm \ref{algo:static_subsets}. It comprises four major steps. We first initialize the current estimate of top-$m$ arms (\textbf{champion set}) $U_0$ to a random set from $\mathcal{S}$ based on utilities estimated by linear surrogate using random $\vec{\theta}$. The stopping criterion is met when there is no ambiguity in discriminating top-$m$ arms in champion set with contenders from challenger set as measured by gap between two most ambiguous arms in these sets. The Gap-index for any pair of arms $i,j$ is computed as $B_t(i,j)=\hat{\mu}_t(i)-\hat{\mu}_t(j)+W_t(i,j)$. Subsequently, if stopping criterion is not met, we re-estimate current champion set of arms (\textbf{Lines 4-9}) by replacing the subcarrier with lowest utility in $U_t$ with the highest utility subcarrier $n_t$ from challenger set $C_t$. Then the challenger set is rotated by considering the set $C_{t-1}$ from previous round and next set of top m' arms from $ U_t^c$ (\textbf{Lines 10-11}) so that the resulting $U_t \cup C_t$ is a high reward set. We then recompute the two most ambiguous arms $b_t \in U_t$ (arm most threatened by some challenger) and $ca_t \in C_t$ (arm with largest chance of beating a top arm) (\textbf{Lines 12-14}). Then we play an arm to observe a reward and update the parameters of the linear surrogate (\textbf{Lines 15-19}). For selecting an arm to play we use the largest variance rule $a^* = \argmin_{a\in N_t\cup U_t} ||x_{b_t} - x_{s_t} ||_{(\hat{V}_{t-1} + x_a x_a^T)^{-1}} $, where arm that minimizes variance between $ca_t$ and $b_t$ is chosen.

\section{Theoretical Analysis}
Following \cite{reda2021top}, we obtain a high probability ($1-\delta$) upper bound on sample complexity of \name{}.

 Let $ \mathcal{S}_\star^{(m)}$ be the true set of top-$m$ arms.
    We define the true gap of an arm 
    $i$ as $\mathcal{G}(i) \triangleq \mu(i)-\mu(m+1)$ if $i \in  \mathcal{S}_\star^{(m)}$, $\mu(m)-\mu(i)$ otherwise ($\mathcal{G}(i) \geq 0$ for any $i \in \ARMS$).

\begin{definition} (Good Gap indices) \\
     \[\mathcal{E} \triangleq \bigcap_{t > 0} \bigcap_{i,j \in \ARMS} \Big(\mu_i-\mu_j \in [-B_t(j,i), B_t(i,j)]\Big),\]
\end{definition}

with $\mathbb{P}(\mathcal{E}) \ge 1-\delta$ which denotes that good  gap indices $B_t(i,j)$ satisfies event $\mathcal{E}$ with probability $\ge$  $1-\delta$.

\begin{theorem}\label{th:upper_bounds_linear_topm}
For \name{}, 
on event $\mathcal{E}$ on which the algorithm is ($\varepsilon, m, \delta$)-PAC, stopping time $\tau_{\delta}$ satisfies $\tau_{\delta} \leq \inf \{u \in \mu^{*+}: u > 1+\HA{}C_{\delta,u}^2 + \mathcal{O}(K)\}$, where, for algorithm  with the largest variance selection rule : 
$\HA{\cA} \triangleq 4\sigma^2{\sum_{a \in \ARMS}} \max \left(\varepsilon,\frac{\varepsilon+\mathcal{G}_{a}}{3} \right)^{-2},$
\end{theorem}

\textbf{Proof Structure for Theorem \ref{th:upper_bounds_linear_topm}}: We build upon the procedure adopted in top-m identification algorithms like LinGapE \cite{xu2018fully} and LinGIFA \cite{reda2021top}. To derive an upper bound on sample complexity, we first state the following Lemma 

\begin{lemma}\label{lemma:m-LinGapE_bound}
On the event $\mathcal{E}$, for all $t > 0$, 
\begin{align*}
    B_t(ca_t, b_t)(t) \leq \min(-({\mathcal{G}(b_t)}  \lor {\mathcal{G}(ca_t)}) +2W_t(b_t,ca_t), 0) \\ +W_t(b_t,ca_t), where a \lor b= \max(a,b)
\end{align*}

\end{lemma}
The intuition behind the lemma is that the estimated worst-case gap is at most the true gap (negative if the ordering is correct) plus a small multiple of the uncertainty. Thus when uncertainty 
$W_t$ is small compared to the true gap, the index becomes negative, which will trigger the stopping rule.

\begin{proof}

    \textbf{Property 1}: For $b_t \in U_t$ and $ca_t \in C_t$
    it holds that $\hat{\mu}_t(b_t)\ge\hat{\mu}_t(ca_t)$.
    Hence, it follows that $B_t(ca_t,b_t) = \hat{\mathcal{G}}_t(ca_t,b_t) + W_t(b_t,ca_t) \le W_t(b_t,ca_t)$ as $\hat{\mathcal{G}}_t(ca_t,b_t)<0$
    
On $\mathcal{E}$, for any $i \notin U_t$ and $j \in U_t$ we have $B_t(i,j)$:
\begin{align*}
 \big(\mu(i)-\mu(j)\big)+\big(\widehat{\mu}_t(i)-\mu(i)-(\widehat{\mu}_t(j)-\mu(j))\big)+W_t(i,j).
\end{align*}
The estimation error is at most $2W_t(i,j)$ (per-arm widths),
\begin{equation}\label{eq:bij-upper}
B_t(i,j)\ \le\ \big(\mu(i)-\mu(j)\big)+2W_t(i,j).
\end{equation}

Applying \eqref{eq:bij-upper} to the maximizing pair $(i,j)=(ca_t,b_t)$ and handling the four exhaustive membership cases of $b_t$ and $ca_t$ with respect to the true top-$m$ set $S_\star^{(m)}$:

\begin{enumerate}
    \item \textbf{$b_t \in S_\star^{(m)}$, $ca_t \notin S_\star^{(m)}$:} Then $\mu(b_t) \ge \mu(m)$, hence $\mu(ca_t)-\mu(b_t) \le - \mathcal{G}(ca_t)$ and $\mu(ca_t)\le \mu(m+1)$ which implies $\mu(ca_t)-\mu(b_t) \le - \mathcal{G}(b_t)$ Substituting into \eqref{eq:bij-upper}
    \begin{align*}
           B_t(ca_t,b_t)\ \le\ -(\mathcal{G}(b_t)\vee\mathcal{G}(ca_t))+2W_t(b_t,ca_t) 
    \end{align*}

    \item \textbf{$b_t \notin S_\star^{(m)}$, $ca_t \in S_\star^{(m)}$:} By Property~1,
    \(
    B_t(ca_t,b_t) \le W_t(b_t,ca_t),
    \)
    which also implies the desired bound.

    \item \textbf{$b_t \notin S_\star^{(m)}$, $ca_t \notin S_\star^{(m)}$:} 
    There exists $b \in S_\star^{(m)}$ that belongs to $C_t$. This is primarily because of how $C_t$ shortlist is sampled in each round in \name{}. As it prioritizes the next top-$m'$  arms with highest means, it captures atleast one arm $\in S_\star^{(m)}$. This implies  $B_t(ca_t,b_t)\ge B_t(b,b_t)$. Chaining the inequality via this $b$ yields \\
    $
    B_t(ca_t,b_t)\ \le\ -(\mathcal{G}(b_t)\vee\mathcal{G}(ca_t)) + 3W_t(b_t,ca_t).
    $

    \item $b_t \in S_\star^{(m)}$, $ca_t \in S_\star^{(m)}$:
    There exists $s \notin S_\star^{(m)}$ in $U_t$ (otherwise the algorithm has already stopped). Repeating the reasoning of case~(i) with $ca$ instead of $ca_t$ gives the same bound up to constants.
\end{enumerate}

Combining the cases and using the fact that 
$B_t(ca_t,b_t)\le W_t(b_t,ca_t)$ (Property~1),
we get the clipped inequality
\begin{align*}
    B_t(ca_t,b_t) \le \min\big(-(\mathcal{G}(b_t)\vee\mathcal{G}(ca_t)) + 2W_t(b_t,ca_t),0\big)\\+W_t(b_t,ca_t)
\end{align*}
Combining above Lemma with stopping rule $B_t(s_t,b_t) \le \epsilon$ following Lemma 8 in \cite{reda2021top} directly yields 
    \[  \NA{a_t}{t} \leq 4\sigma^2 C_{\delta,t}^2\max \left( \varepsilon, \frac{\varepsilon+\mathcal{G}_{a_t}}{3} \right)^{-2}\]
where $\NA{a_t}{t}$ is the number of times arms $a$ is sampled.This is equivalent  to $\HA{\cA}$ in Theorem 1. Hence, maximum number of samplings on event $\mathcal{E}$ is upper-bound by $\inf_{u \in \mathbb{R}^{*+}} \{u > 1+\HA{\text{LinGIFA}}C_{\delta,u}^2\}$, where $\HA{\text{LinGIFA}} \triangleq 4\sigma^2\sum_{a \in \ARMS} \max \left( \varepsilon, \frac{\varepsilon+\mathcal{G}_{a}}{3} \right)^{-2}$.
\end{proof}

\begin{figure*}[hbt!]
\small
\centering 
\begin{subfigure}{0.30\linewidth}
\begin{tikzpicture}[every mark/.append style={mark size=1.2pt}]
    \begin{axis}[
        boxplot/draw direction = y,
            width=0.95\linewidth,
            height=4.3cm,
            ylabel style = {font= \tiny},
            ylabel={Average runtime (in seconds)},
            xticklabel style={rotate=0},
            xtick = {1, 2, 3,4},
            yticklabel style = {font=\boldmath \tiny, xshift=0.5ex},
        xticklabel style ={font=\small , yshift=0.2ex},
            ymajorgrids,
		xticklabels = {\name{}, LG, LUG, LGI},
            area legend,
            legend cell align={left},
            legend style={
                    cells={align=left},
                },        ]

        \addplot+[
            boxplot={draw position=1},
        ] table[
                y=time_taken,
                col sep=comma,
            ] {plots/K_40/CASE_Heuristic_greedy_time_K=40.csv};
        \addplot+[
            boxplot={draw position=2},
        ] table[
                y=time_taken,
                col sep=comma,
            ] {plots/K_40/LinGapE_Heuristic_greedy_time_K=40.csv};
        \addplot+[
            boxplot={draw position=3},
        ] table[
                y=time_taken,
                col sep=comma,
            ] {plots/K_40/LinUGapE_Heuristic_time_K=40.csv};
    \addplot+[
            boxplot={draw position=4},
        ] table[
                y=time_taken,
                col sep=comma,
            ] {plots/K_40/LinGIFA_Heuristic_greedy_time_K=40.csv};

    \end{axis}
\end{tikzpicture}
\caption{ $K=40$, $m=12$}
\end{subfigure}
\hspace{-3em}
\begin{subfigure}{0.35\linewidth}
\begin{tikzpicture}[every mark/.append style={mark size=1.2pt}]
    \begin{axis}[
        boxplot/draw direction = y,
            width=0.95\linewidth,
            height=4.3cm,
            ylabel style = {font= \tiny},
            ylabel={Average runtime (seconds)},
            xticklabel style={rotate=0},
            xtick = {1, 2, 3,4},
            yticklabel style = {font=\boldmath \tiny, xshift=0.5ex},
        xticklabel style ={font=\small , yshift=0.2ex},
            ymajorgrids,
		xticklabels = {\name{}, LG, LUG, LGI},
            area legend,
            legend cell align={left},
            legend style={
                    cells={align=left},
                },        ]

        \addplot+[
            boxplot={draw position=1},
        ] table[
                y=time_taken,
                col sep=comma,
            ] {plots/K_40/CASE_Heuristic_greedy_time_K=100.csv};
        \addplot+[
            boxplot={draw position=2},
        ] table[
                y=time_taken,
                col sep=comma,
            ] {plots/K_40/LinGapE_Heuristic_greedy_time_K=100.csv};
        \addplot+[
            boxplot={draw position=3},
        ] table[
                y=time_taken,
                col sep=comma,
            ] {plots/K_40/LinUGapE_Heuristic_time_K=100.csv};
    \addplot+[
            boxplot={draw position=4},
        ] table[
                y=time_taken,
                col sep=comma,
            ] {plots/K_40/LinGIFA_Heuristic_greedy_time_K=100.csv};

    \end{axis}
\end{tikzpicture}
\caption{$K=100$, $m=12$}
\end{subfigure}
\hspace{-4em}
\begin{subfigure}{0.30\linewidth}
\begin{tikzpicture}[every mark/.append style={mark size=1.2pt}]
    \begin{axis}[
        boxplot/draw direction = y,
            width=0.95\linewidth,
            height=4.3cm,
            ylabel style = {font= \tiny},
            ylabel={Average runtime (seconds)},
            xticklabel style={rotate=0},
            xtick = {1, 2, 3,4},
            yticklabel style = {font=\boldmath \tiny, xshift=0.5ex},
        xticklabel style ={font=\small , yshift=0.2ex},
            ymajorgrids,
		xticklabels = {\name{}, LG, LUG, LGI},
            area legend,
            legend cell align={left},
            legend style={
                    cells={align=left},
                },        ]

        \addplot+[
            boxplot={draw position=1},
        ] table[
                y=time_taken,
                col sep=comma,
            ] {plots/K_40/CASE_Heuristic_greedy_time_K=600.csv};
        \addplot+[
            boxplot={draw position=2},
        ] table[
                y=time_taken,
                col sep=comma,
            ] {plots/K_40/LinGapE_Heuristic_greedy_time_K=600.csv};
        \addplot+[
            boxplot={draw position=3},
        ] table[
                y=time_taken,
                col sep=comma,
            ] {plots/K_40/LinUGapE_Heuristic_time_K=600.csv};
    \addplot+[
            boxplot={draw position=4},
        ] table[
                y=time_taken,
                col sep=comma,
            ] {plots/K_40/LinGIFA_Heuristic_greedy_time_K=600.csv};

    \end{axis}
\end{tikzpicture}
\caption{ $K=600$, $m=12$}
\end{subfigure}
\caption{Top-$m$ arm identification runtime comparison between \name{}, LinGIFA (LGI), LinUGapE (LUG) and LinGapE (LG). }
\label{fig:comparison_time_20}
\end{figure*}
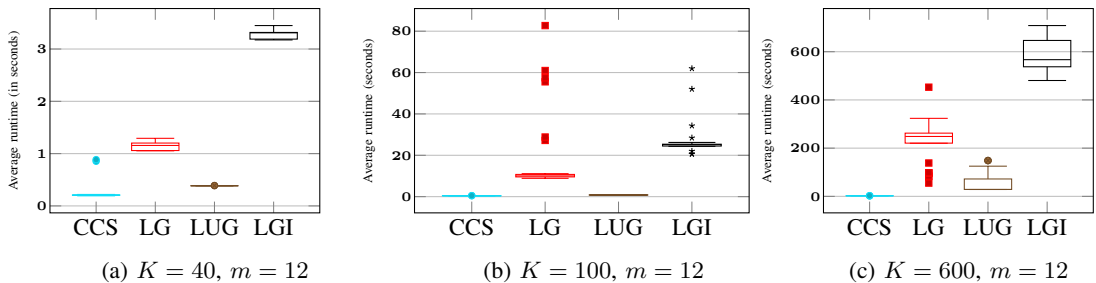

\begin{figure}[hbt!]
\small
 
\begin{subfigure}{0.50\linewidth}
\begin{tikzpicture}
\begin{axis}[
    ylabel style = {font= \tiny},
    xlabel style = {font= \small},
    xlabel= $|C_t|$,
    xticklabel style={font=\small},
    yticklabel style={font=\boldmath \tiny},
    ylabel=Correctness (\%),
    height=4.1cm,
    legend style={
font= \tiny,at={(0.7,0.3)},anchor=south east},
    width=5.5cm,
    xmin=5, xmax=300,
    ymin=0, ymax=120]
    ytick={0,0.1,0.2,0.3}
    xtick={1,10,25,50,75,300}
\addplot[smooth,mark=*,red,        
            error bars/.cd,
                y dir=both,
                y explicit,] plot      
    coordinates {
    (10,30.00) +-(10,13.67)
    (50,75.3) +-(50, 13.98)
    (100,93.50) +-(100,6.58)
    (150,98.33)+-(150, 4.55)
    (200,99.66)+-(200, 1.64)
    (250,100)
    (300,100)};
\addlegendentry{\name{}}

\end{axis}
    \end{tikzpicture}
\vspace{-1em}
\caption{Avg. correctness}
\end{subfigure}
\hspace{1em}
\begin{subfigure}{0.40\linewidth}
\begin{tikzpicture}
\begin{axis}[
    ylabel style = {font= \tiny},
    xlabel style = {font= \small},
    xlabel= $|C_t|$,
    xticklabel style={font=\small},
    yticklabel style={font=\boldmath \tiny},
    ylabel=Time (in seconds),
    height=4.1cm,
    legend style={
                    font= \tiny,
                },
    width=4.5cm,
    xmin=5, xmax=300,
    ymin=0, ymax=5]
    ytick={0,0.1,0.2,0.3}
    xtick={1,10,25,50,75,300}
\addplot[smooth,mark=*,red, 
            error bars/.cd,
                y dir=both,
                y explicit,] plot      
    coordinates {
    (10,0.67) +-(10,0.11)
    (50,1.05) +-(50, 0.10)
    (100,1.50) +-(100,0.11)
    (150,2.08)+-(150, 0.23)
    (200,2.50)+-(200, 0.15)
    (250,3.04)+-(250, 0.24)
    (300,3.50)+-(300, 0.19)};
\addlegendentry{\name{}}

\end{axis}
    \end{tikzpicture}
\vspace{-1em}
\caption{Average runtime }
\end{subfigure}
\caption{Variation in correctness and latency of \name{} with $|C_t|$. }
\label{fig:correctness_vs_ndas}
 \vspace{-3mm}
\end{figure}
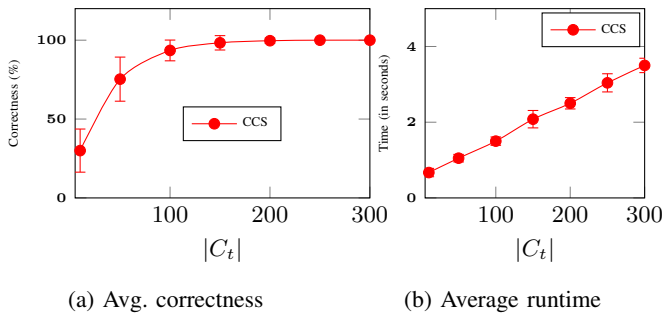

\vspace{-1em}
\section{Simulation Results}

\subsection{Experimental Setup}
\label{sec:exp-setup}
We evaluate \name{} in an OFDM downlink where each arm is a subcarrier and each action selects one arm, isolating per-tone exploration. We use LTE/5G numerology with \(\Delta f=15\,\mathrm{kHz}\) and carrier bandwidths \(\{10,1.5,0.6\}\,\mathrm{MHz}\), yielding \(K\in\{600,100,40\}\) active tones \cite{goldsmith2005wireless, morais20245g}. Scheduling uses \(m=12\) tones (one Resource Block (RB): \(12\times15\,\mathrm{kHz}=180\,\mathrm{kHz}\)) \cite{morais20245g, salem2010opportunities}. Transmit power is fixed per tone \(P_{\text{tone}}\approx2.22\,\mathrm{dBm}\ (\approx1.67\,\mathrm{mW})\), so \(P_{\mathrm{tot}}[\mathrm{dBm}]=2.22+10\log_{10}K\) (e.g., \(\approx30,22.2,18.2\,\mathrm{dBm}\) for \(K=600,100,40\)). With \(\mathrm{NF}=5\,\mathrm{dB}\) and thermal noise \(-174\,\mathrm{dBm/Hz}\), the per-tone noise is \(\sigma^2[\mathrm{dBm}]=-174+10\log_{10}(\Delta f)+\mathrm{NF}\approx-127.24\,\mathrm{dBm}\) (\(\sigma^2\approx1.89\times10^{-16}\,\mathrm{W}\)). Pathloss sets the median SNR (\(\mathrm{PL}=-120\,\mathrm{dB}\)); small-scale fading is Rayleigh with \(h_i\sim\mathcal{CN}(0,1)\). Thus \(\gamma_i=\frac{|h_i|^2P\,10^{\mathrm{PL}/10}}{\sigma^2}\) and the reward is \(Y_t=\log_2(1+\gamma_i)+\eta_t\) with sub-Gaussian \(\eta_t\). We use a 20th-order Taylor linear surrogate (features \(X_i\in\mathbb{R}^{20}\)). CPU-only experiments run on 4 Alvis nodes (2\(\times\)32-core Xeon Gold 8358, 512\,GiB RAM) \cite{c3se_alvis_cpu_only}. Unless noted, we run 50 trials per setting with \(m=12\), \(\delta=0.05\), \(\varepsilon=10^{-15}\), \(K\in\{40,100,600\}\), \(d=20\), noise proxy \(\sigma=1\), and challenger sizes \(|C_t|\in\{10,20,200\}\) for \(K=\{40,100,600\}\).

\noindent Our Code: \href{https://github.com/mohsen1amiri/CCS_in_OFDM.git}{ \small \texttt{github.com/mohsen1amiri/CCS\_in\_OFDM}}

\subsection{Baselines}
We evaluate the performance of the proposed challenger-based selective exploration algorithm in the context of OFDM subcarrier selection. We compare against representative baseline methods and report results in terms of efficiency and accuracy of identifying the top-$m$ subcarriers:
\begin{itemize}
    \item \textbf{Exhaustive Evaluation (Oracle):} All $\binom{K}{m}$ subsets are enumerated and evaluated to identify the true top-$m$ arms. This method provides an upper bound on achievable accuracy but is computationally intractable except for small $K$.

    \item \textbf{LinGapE} \cite{xu2018fully}: An efficient adaptive sampling algorithm that was the first to propose use of paired indices.

        \item \textbf{LinUGapE} \cite{LinUGapE}: An extension of UGapE \cite{LinUGapE} to top-$m$ setting through sequential runs  and selects two arms with the largest upper confidence bound on the gap between their means and samples arms that help discriminate between this pair with a local greedy approach.

            \item \textbf{LinGIFA} \cite{reda2021top}: A  gap-index algorithm which instead of a local greedy approach it considers current estimates of all top-$m$ and non top-$m$ arms and allocates samples to minimize uncertainty across the set. However, this requires comparing all pairs of arms each round leading to large number of gap-index computations and latency.

\end{itemize}

\begin{table}[h]
\centering

\caption{Number of comparisons for gap computations (average $\pm$ std) across number of subcarriers (K)}
\label{tab:comparisons_ave_std}
\setlength{\tabcolsep}{3pt} 
\footnotesize               
\resizebox{\columnwidth}{!}{%
\begin{tabular}{llll}
\toprule
{} & K=40 & K=100 & K=600 \\
\midrule
\name{}      & \textbf{3,940} $\pm$ 23   & \textbf{8,076} $\pm$ 61   & \textbf{56,533} $\pm$ 441 \\
LinGIFA   & 3,299,145 $\pm$ 23,840      & 61,828,686 $\pm$ 1,697,091   & 9,301,157,253 $\pm$ 249,697,551 \\
LinGapE   & 1,070,350 $\pm$ 16,754      & 24,602,769 $\pm$ 48,4424    & 3,525,220,180 $\pm$ 110,965,182 \\
LinUGapE  & 384,156 $\pm$ 0           & 2,000,112 $\pm$ 0          & 432,000,612 $\pm$ 0 \\
\bottomrule
\end{tabular}%
}
\end{table}


\subsection{Results}

\subsubsection{Efficiency}
Table~\ref{tab:comparisons_ave_std} summarizes the average number of pairwise gap-index comparisons required to stop.  Across all three problem sizes, \name{} reduces comparisons by orders of magnitude relative to all baselines. Concretely, for \(K{=}40\), \name{} averages \(3.94{\times}10^{3}\) comparisons versus \(1.07{\times}10^{6}\) (LinGapE), \(3.30{\times}10^{6}\) (LinGIFA), and \(3.84{\times}10^{5}\) (LinUGapE), i.e., \(\approx\!272{\times}\), \(837{\times}\), and \(97{\times}\) fewer, respectively. For \(K{=}100\), the savings grow to \(\approx\!3.0{\times}10^{3}{\times}\) (LinGapE), \(7.7{\times}10^{3}{\times}\) (LinGIFA), and \(2.5{\times}10^{2}{\times}\) (LinUGapE). At the largest scale \(K{=}600\), \name{} requires only \(5.65{\times}10^{4}\) comparisons on average, versus \(3.53{\times}10^{9}\) (LinGapE), \(9.30{\times}10^{9}\) (LinGIFA), and \(4.32{\times}10^{8}\) (LinUGapE), corresponding to \(\approx\!6.2{\times}10^{4}{\times}\), \(1.6{\times}10^{5}{\times}\), and \(7.6{\times}10^{3}{\times}\) reductions, respectively. These gains are consistent with the design of \name{}: by confining attention to a small champion set and a rotating shortlist of challengers, the algorithm avoids uninformative pairwise tests.

\subsubsection{Runtime}
The dramatic reduction in comparisons translates into equally large runtime savings. The boxplots in Fig.~\ref{fig:comparison_time_20} show that \name{} completes substantially faster than all baselines for all \(K\), with the runtime gap widening as \(K\) increases. For \(K{=}40\) (Fig.~\ref{fig:comparison_time_20}a), \name{} already dominates; by \(K{=}100\) and \(K{=}600\) (Fig.~\ref{fig:comparison_time_20}b--c), baseline runtimes balloon by two--three orders of magnitude, whereas \name{} remains near the floor, reflecting its constant-size shortlists and targeted sampling. Overall, this yields speedups of \textbf{100$\times$}, \textbf{15$\times$}, and \textbf{300$\times$} over LinGapE~\cite{xu2017fullyadaptivealgorithmpure}, LinUGapE~\cite{LinUGapE}, and LinGIFA~\cite{reda2021top}, respectively, while maintaining near-perfect top-\(m\) identification accuracy.

\subsubsection{Convergence and stability}
Fig.~2 illustrates the evolution of identification accuracy (``Correctness'') and cumulative time over the set of challenger arms \(|C_t|\) for \name{}. Correctness climbs rapidly and saturates near 100\% within a few hundred challenger arms, while elapsed time grows roughly linearly with rounds and remains small, consistent with the bounded per-round work induced by champion, challenger selection.

\subsubsection{Number of arm pulls}
For an \emph{exhaustive evaluation} we must consider all \(m\)-subsets, i.e., \(\binom{K}{m}\).
For \(K\in\{40,100,600\}\) and \(m=12\), this is approximately
\(\binom{40}{12}\approx 5.58\times 10^{9}\), \(\binom{100}{12}\approx 1.05\times 10^{15}\),
and \(\binom{600}{12}\approx 4.06\times 10^{24}\) subsets, respectively.
Because rewards are noisy (due to SNR fluctuations), each arm must be sampled \(M\) times to estimate its mean; evaluating one subset therefore costs \(mM\) pulls.
The total number of pulls is thus $T \;=\; m\,M\,\binom{K}{m}$,
which is prohibitively large, making exhaustive search infeasible. However, the proposed MAB-based approach \name{} and other MAB baselines reported here are much sample efficient, requiring fewer arm (subcarrier) evaluations to sample rewards. Among the MAB approaches, we observe that LinGapE and our approach \name{} have similar sample complexity (arm pulls), closely followed by LinGIFA, and LinUGapE has the highest number of arm pulls. For instance, in $K=600$ setup, \name{} and LinGapE have an average of 17.46 arm pulls across simulations followed by LinGIFA with an average of 22.12. Finally, LinUGapE has the highest number of arm evaluations, amounting to 601 arm pulls. It is to be noted that \name{} has the lowest latency with an average runtime of 2 seconds, whereas even LinGapE takes 210 seconds for top-$m$ arm selection as shown in Fig. \ref{fig:comparison_time_20}.

\subsection{Discussion}
\label{subsec:discussion}

Across all scales \(K\in\{40,100,600\}\), \name{} reduces both gap-index comparisons and runtime by one to three orders of magnitude compared with representative linear baselines. For \(K{=}600\), Table~\ref{tab:comparisons_ave_std} reports factors of \(\approx 6.2\times 10^{4}\), \(1.6\times 10^{5}\), and \(7.6\times 10^{3}\) fewer comparisons than LinGapE, LinGIFA, and LinUGapE, respectively, mirroring the pronounced runtime gap in Fig.~\ref{fig:comparison_time_20}. Even at smaller \(K\), \name{} already dominates, as shown in Fig.~\ref{fig:comparison_time_20}. 

Fig.~\ref{fig:correctness_vs_ndas} highlights the trade-off controlled by \(|C_t|\). Increasing \(|C_t|\) improves the chance that the \emph{true} adversary to the current champion is present in the shortlist, which accelerates elimination and pushes correctness to \(\approx 100\%\); however, the runtime grows roughly linearly with \(|C_t|\). In our runs, correctness saturates after a few hundred challenger size, beyond which the marginal benefit is small relative to the added latency. 

In a realistic setting with \(K{=}600\) subcarriers, \name{} identifies the top-\(m\) arms in about \(2\text{--}4\) seconds. With a highly optimized implementation or fast hardware, this could plausibly be pushed to the sub-millisecond range, a capability the other baselines do not exhibit. These gains come without observable loss in identification accuracy: the correctness curves saturate near \(100\%\) once the challenger list is moderately sized (Fig.~\ref{fig:correctness_vs_ndas}). Moreover, in applications where near-perfect accuracy is unnecessary and speed is paramount, reducing the challenger set size \(|C_t|\) can further accelerate computation (up to \(\sim 4\times\) faster) while maintaining acceptable accuracy (Fig.~\ref{fig:correctness_vs_ndas}). 

Lastly, in terms of sampling budget, \name{} matches LinGapE in arm pulls (average \(\approx 17.5\) at \(K{=}600\)), stays below LinGIFA (\(\approx 22.1\)), and is well over an order of magnitude below LinUGapE (\(\approx 601\)); thus, the speedups do not rely on increased sampling and require only estimating the corresponding SNRs of the few pulled arms (subcarriers).

The main driver is \emph{where} computation and samples are spent. Algorithms such as LinGapE/LinGIFA compute indices over broad swathes of the action set to find the most ambiguous pair each round. In contrast, \name{} confines attention to the challenger shortlist $C_t$ of plausible adversaries. This design reduces both the search cost (forming at most $O(m|C_t|)$ pairwise indices per round rather than $\Omega(K^2)$ in the worst case) and the sample cost (since only the most informative champion--challenger comparisons are pulled). As $K$ grows while $|C_t|$ is kept moderate, the gains in both comparisons and runtime widens, as observed in Table~\ref{tab:comparisons_ave_std} and Fig.~\ref{fig:comparison_time_20}.

Our analysis assumes a linear utility in features derived from noisy SNR, instantiated here via a 20-term Taylor surrogate of $\log_2(1+\gamma)$ (Sec.~\ref{sec:system_model}). While this captures the local curvature well over mid-SNRs, very high or very low SNRs can cause numerical stress for high-order polynomials. However, here the 20-term shows sufficient performance over different scenarios. Nonetheless, for creating the feature vectors, the noisy SNRs suffice, and we can estimate the reward function only using them. Therefore, we are estimating the reward function (data rate) for each arm (subcarrier) only using the noisy SNR estimation of each subcarrier. Notably, we showed experimentally that even when SNR is noisy with 1 dB standard deviation, \name{} still can find the top-m arms (subcarriers).

By design, our simulator isolates the exploration challenge to per-subcarrier decisions: each arm is a single subcarrier, and we ultimately identify the best $m$ subcarriers under an \emph{additive} utility ($U(S)=\sum_{i\in S}R_i$). Under this model, selecting the top-$m$ arms individually is equivalent to selecting the best size-$m$ subset. In operational OFDM stacks, additional constraints may break this equivalence—e.g., RB contiguity, per-user grouping, fairness across users/slices, or power-coupling across subcarriers. In such settings, \name{} can still be applied by redefining ``arms'' as feasible primitives (e.g., RBs or small RB bundles) and letting the champion set hold $m$ such primitives; the shortlist then contains only feasible challengers that respect contiguity and policy constraints. Exploring this constrained variant is a natural next step.


\section{Conclusion}
We cast top-(m) OFDM downlink subcarrier selection as a pure-exploration stochastic linear bandit problem and propose \name{}, which maintains champion/challenger shortlists and compares them via a gap index. Using a linear SNR($\to$)rate surrogate, \name{} concentrates measurements and achieves orders-of-magnitude fewer gap-index computations and lower runtime than LinGapE/LinUGapE/LinGIFA with near-identical accuracy. It offers a speed-accuracy knob (challenger set size - $|C_t|$) and (($\varepsilon,m,\delta$))-PAC guarantees, making online, measurement-efficient selection practical under tight latency. The future work includes RB contiguity, multi-user MIMO/fairness, frequency correlation/power coupling, adaptive shortlists, and hardware-in-the-loop evaluation.

\section*{Acknowledgment}

This work was supported by the Swedish Research Council (2024-04058), Vinnova; compute from NAISS at C3SE, partly funded by the Swedish Research Council (2022-06725).

\bibliographystyle{IEEEtran}
\bibliography{references}

\end{document}